%% file: main.tex
\documentclass{article}
\usepackage{times}
\usepackage{multicol}
\usepackage{multirow}
\usepackage{booktabs}
\usepackage{threeparttable}
\makeatletter
\newcommand{\setteaser}{\def\@captype{figure}}
\makeatother
\usepackage{graphicx}
\usepackage{wrapfig}
\usepackage{color}
\usepackage{amsmath, amssymb, amsthm}
\graphicspath{{figures/}}
\usepackage[bookmarks=true]{hyperref}
\usepackage[final]{corl_2026} 

\title{CMP: Robust Whole-Body Tracking for Loco-Manipulation via Competence Manifold Projection}

\author{
  Ziyang Cheng$^{1,2,3}$ \quad
  Haoyu Wei$^{1,2,3}$ \quad
  Hang Yin$^{1,2,3}$ \quad
  Xiuwei Xu$^{1,2,3}$ \\
  \textbf{Bingyao Yu}$^{1,2,3}$ \quad
  \textbf{Jie Zhou}$^{1,2,3}$ \quad 
  \textbf{Jiwen Lu}$^{1,2,3,*}$ \\
  $^1$Department of Automation, Tsinghua University \\
  $^2$Beijing Key Laboratory of Embodied Intelligence Systems \\
  $^3$Institute for Embodied Intelligence and Robotics, Tsinghua University \\
  $^*$Corresponding author \\
}

\hypersetup{
  pdftitle={CMP: Robust Whole-Body Tracking for Loco-Manipulation via Competence Manifold Projection},
  pdfauthor={Ziyang Cheng, Haoyu Wei, Hang Yin, Xiuwei Xu, Bingyao Yu, Jie Zhou, Jiwen Lu}
}

\begin{document}
\maketitle
\begin{center}
    \vspace{-25pt}
    \centering
    \setteaser
    \includegraphics[width=1\textwidth]{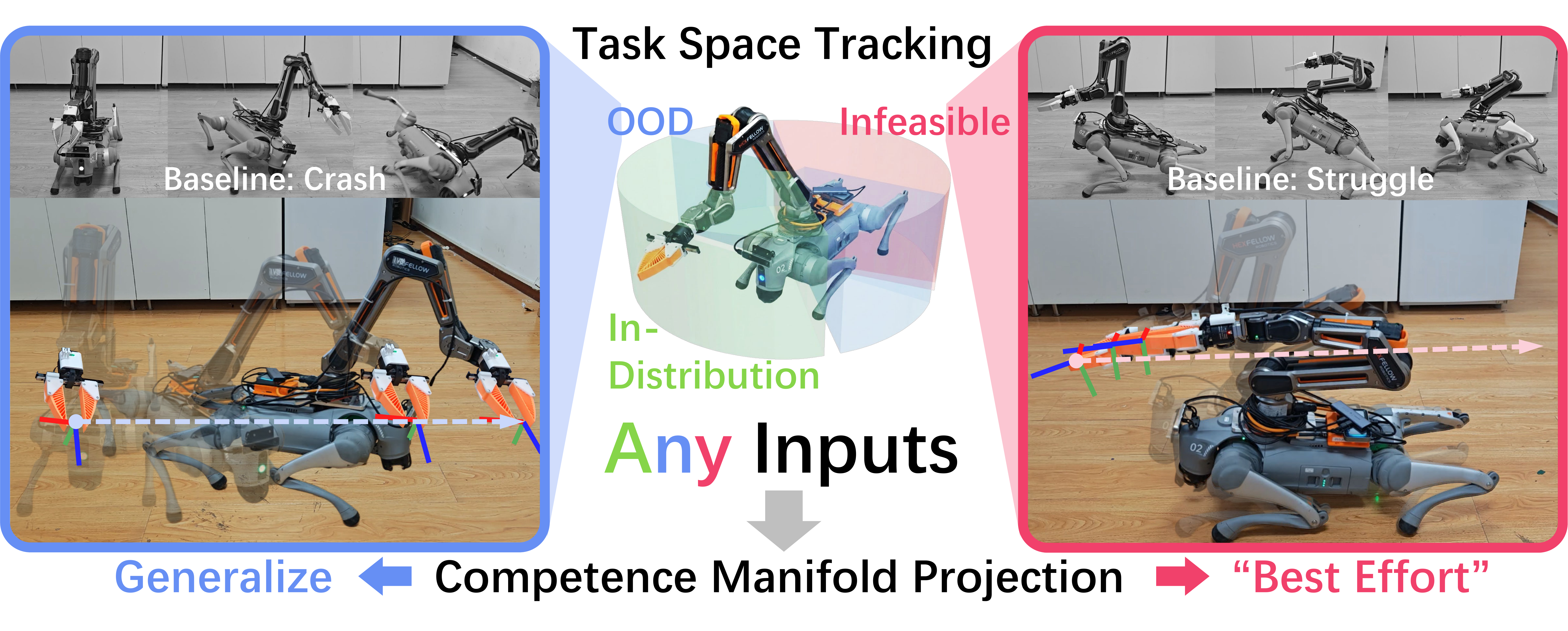}
    \vspace{-25pt}
    \caption{\textbf{Robust Whole-Body Tracking via Competence Manifold Projection (CMP).}
    The framework processes arbitrary task-space inputs for a legged manipulator, conceptually divided into In-Distribution (ID, green), Out-of-Distribution (OOD, blue), and Infeasible (red) regions.
    By projecting inputs onto a learned safety manifold:
    (Left) In OOD scenarios (e.g., a sideways push), the robot demonstrates emergent generalization, successfully tracking OOD trajectories by adhering to its competence boundary.
    (Right) For infeasible commands (e.g., a backward push far beyond model competence), the system exhibits ``best-effort'' behavior, safely interacting as close to the target as possible without crossing into unsafe states.}
    \label{fig:teaser}
\end{center}
\vspace{-6pt}


\input{sections/00_abstract}



\input{sections/01_introduction}
\input{sections/02_related_work}
\input{sections/03_method}
\input{sections/04_experiments}
\input{sections/05_conclusion}
\input{sections/06_limitations}


\clearpage
\input{sections/07_acknowledgments}


\bibliography{references}  

\input{sections/08_appendix}

\end{document}

%% file: sections/00_abstract.tex
\begin{abstract}
While decoupled control schemes for legged mobile manipulators have shown robustness, learning holistic whole-body control policies for tracking global end-effector poses remains fragile against Out-of-Distribution (OOD) inputs induced by sensor noise or infeasible user commands. To improve robustness against these perturbations without sacrificing task performance and continuity, we propose \emph{Competence Manifold Projection} (CMP).
Specifically, we utilize a \emph{Frame-Wise Safety Scheme} that transforms the infinite-horizon safety constraint into a computationally efficient single-step manifold inclusion.
To instantiate this competence manifold, we employ a \emph{Lower-Bounded Safety Estimator} that distinguishes unmastered intentions from the training distribution.
We then introduce an \emph{Isomorphic Latent Space} (ILS) that aligns manifold geometry with safety probability, enabling efficient $\mathcal{O}(1)$ seamless defense against arbitrary OOD intents.
Experiments demonstrate that CMP achieves up to a \emph{10-fold} survival rate improvement in typical OOD scenarios where baselines suffer catastrophic failure, incurring \emph{under 10\%} tracking degradation. Notably, the system exhibits emergent ``best-effort'' generalization behaviors to progressively accomplish OOD goals by adhering to the competence boundaries.
Project website: \href{https://shepherd1226.github.io/CMP/}{https://shepherd1226.github.io/CMP/}.
\end{abstract}

%% file: sections/01_introduction.tex
\section{Introduction}

Legged mobile manipulators represent a versatile class of robots capable of traversing unstructured environments while performing complex manipulation tasks.
To fully exploit their potential, recent paradigms, exemplified by UMI-on-Legs~\cite{Huy2024UMIonLegs}, have shifted towards global end-effector tracking in the task space.
This approach offers seamless compatibility with cross-embodiment data collection pipelines like UMI~\cite{chi2024umi} and FastUMI~\cite{Zhaxizhuoma2025FastUMI}, end-effector-centric teleoperation interfaces, and high-level planning policies, effectively mobilizing whole-body coordination.
By receiving target poses in the global frame, localizing the current end-effector via Visual-Inertial Odometry (VIO), and computing the relative target pose for policy execution, this framework guarantees native spatial translation generalization for the Whole-Body Control (WBC) policy while stably tracking global targets.

However, this unified coordination introduces significant reliability challenges.
The learned policies are typically only competent for specific trajectories within the training dataset or limited spatial distributions (e.g., the natural manipulation workspace located in front of the robot~\cite{Huy2024UMIonLegs, liuMLMLearningMultitask2025, Jiang2025Learning}).
Here, \emph{competence} denotes the policy's ability to operate safely.
When inputs exceed this range---falling Out-of-Distribution (OOD, referring to commands outside the policy's competence distribution rather than the training distribution, as detailed in Section~\ref{subsec:method_overview})---due to VIO drift, teleoperation latency, or infeasible user commands, the policy often manifests unpredictable and physically unsafe behaviors~\cite{zhangSafeLearningContactRich2025, xingRobustSecureEmbodied2025a}.
While fully or partially decoupled WBC architectures~\cite{Ma2022Combining, Liu2024VisualWholeBody, Yokoyama2024ASC, panRoboDuetLearningCooperative2025, Fu2023DeepWholeBody} offer robustness, they inherently limit whole-body synergy or compatibility with task-space paradigms~\cite{Huy2024UMIonLegs, chi2024umi, Zhaxizhuoma2025FastUMI}.
Furthermore, existing safety solutions fail to adequately address three critical issues in this context.
First, inference-time policy steering methods~\cite{sunLatentPolicyBarrier2025, janner2022planning, ajay2022conditional, zhou2024diffusion, sun2024force, qi2025strengthening, nakamura2025generalizing} face severe real-time constraints, often being too computationally expensive for high-frequency and highly dynamic WBC loops.
Second, standard OOD detectors~\cite{xu2025can, he2024rediffuser, ganaiRealTimeOutofDistributionFailure2025} confuse ``Out-of-Training-Distribution'' with ``Out-of-Safe-Distribution'', thus failing to identify actions that lie within the training distribution but were never successfully mastered.
Finally, traditional reactive triggers~\cite{ma2023learning, yang2020multi, mengSafeFallLearningProtective2025} or emergency stops~\cite{römerFailurePredictionRuntime2025} inevitably disrupt task continuity, lacking a ``best-effort'' mechanism to smoothly degrade performance while enforcing safety.
Representative methods are compared in Table~\ref{tab:comparison}.

\begin{wraptable}[6]{r}{0.65\textwidth}
    \centering
    \vspace{-20pt}
    \caption{Comparison of representative approaches.}
    \label{tab:comparison}
    \resizebox{\linewidth}{!}{
    \begin{tabular}{lcccc}
        \toprule
        \textbf{Method} & \textbf{Holistic} & \textbf{Competence-Aware} & $\boldsymbol{\mathcal{O}(1)}$ & \textbf{Best-Effort} \\
        \midrule
        Decoupled WBC~\cite{Ma2022Combining, Liu2024VisualWholeBody}                                    & $\times$   & $\times$   & $-$ & $-$   \\
        End-to-end WBC~\cite{Huy2024UMIonLegs, liuMLMLearningMultitask2025, Jiang2025Learning}          & \checkmark & $\times$   & $-$ & $-$   \\
        OOD Detectors~\cite{ganaiRealTimeOutofDistributionFailure2025, römerFailurePredictionRuntime2025} & $-$        & \checkmark   & $\times$ & $\times$   \\
        Inf.-time Steering~\cite{sunLatentPolicyBarrier2025, janner2022planning}                         & $-$        & \checkmark   & $\times$   & \checkmark   \\
        \textbf{CMP (Ours)}                                                                              & \checkmark & \checkmark & \checkmark & \checkmark \\
        \bottomrule
    \end{tabular}
    }
\end{wraptable}

To address these issues, we propose \emph{Competence Manifold Projection} (CMP). 
First, we establish a \emph{Frame-Wise Safety Scheme} to decouple temporal safety, effectively transforming the intractable infinite-horizon constraint into a verifiable single-step manifold inclusion condition. 
To ground this manifold, we then develop a \emph{Lower-Bounded Safety Estimator} that distinguishes mastered from unmastered behaviors, thereby resolving safety boundary ambiguity. 
Finally, these components are unified via the \emph{Isomorphic Latent Space} (ILS), which aligns safety probability with manifold geometry to enable efficient $\mathcal{O}(1)$ projection with a run-time tunable trade-off between tracking precision and safety. 
This integrated pipeline ensures a seamless continuum of control: it preserves full performance for safe inputs while autonomously degrading to the closest feasible behavior for unsafe ones, enabling ``best-effort'' tracking along the capability boundary. 
Experimental evaluations show that CMP enhances survival rates by up to a factor of 10 across typical OOD scenarios in both simulation and real-world deployments, while strictly bounding in-distribution tracking degradation to under 10\%.

%% file: sections/02_related_work.tex
\section{Related Work}

\subsection{Whole-Body Control for Legged Manipulators}

Control architectures for legged manipulators have evolved to navigate the trade-off between modular stability and whole-body synergy. Early approaches explicitly decouple locomotion from manipulation using Model Predictive Control (MPC)~\cite{Bellicoso2019ALMA, Sleiman2023Versatile} or hierarchical learning~\cite{Ma2022Combining, Liu2024VisualWholeBody}, typically treating the arm as a disturbance or solving for motions sequentially. While this modularity accommodates robust navigation planners~\cite{dudzik2020robust, buchanan2021perceptive, hoeller2021learning} and simplifies sim-to-real transfer~\cite{lee2020learning, guHumanoidGymReinforcementLearning2024}, it inherently restricts the reachable workspace by neglecting whole-body momentum~\cite{Yokoyama2024ASC}. To recover coordination, unified policies often define objectives in the floating-base frame~\cite{panRoboDuetLearningCooperative2025, Fu2023DeepWholeBody}. However, this formulation introduces high-frequency jitter transmission~\cite{Fu2023DeepWholeBody} and suffers from interface mismatch with large-scale global-pose datasets like UMI~\cite{chi2024umi, Zhaxizhuoma2025FastUMI}.

Recent end-to-end frameworks~\cite{Huy2024UMIonLegs, liuMLMLearningMultitask2025, Jiang2025Learning} directly track global end-effector poses to leverage onboard state estimation for agile maneuvers. Yet, without intrinsic competence awareness, these policies remain notoriously fragile against OOD commands induced by sensor noise or infeasible user inputs~\cite{zhangSafeLearningContactRich2025, xingRobustSecureEmbodied2025a}. Addressing this, our framework maintains the dynamic advantages of holistic whole-body coordination and seamless compatibility with mainstream global-pose interfaces, while introducing a latent projection layer to robustify the policy against arbitrary upper-level perturbations.

\subsection{Physical Safety and OOD Handling}

Ensuring reliability in learning-based control requires addressing both physical constraints and distributional shifts efficiently~\cite{brunke2022safe}. Traditional mechanisms often rely on ``break-then-fix'' reactive recoveries~\cite{ma2023learning, yang2020multi, huangLearningHumanoidStandingup2025, mengSafeFallLearningProtective2025, kumarLearningControlPolicies2022}, which intervene only after stability is compromised, or impose conservative constraints that hinder exploration~\cite{OpenAI2019SafeRL, tessler2018reward, dalal2018safe, cheng2019end}. While predictive shielding via Control Barrier Functions (CBFs)~\cite{benaGeometryAwarePredictiveSafety2025, yangSHIELDSafetyHumanoids2025, ames2019control, emamSafeReinforcementLearning2022, wang2025end} or Hamilton-Jacobi reachability~\cite{hsu2021safety, bansal2017hamilton, thananjeyan2021recovery} offer foresight, they necessitate manual, task-specific safety function design and struggle to scale. Conversely, data-driven OOD detectors~\cite{xu2025can, he2024rediffuser, ganaiRealTimeOutofDistributionFailure2025} and failure monitors~\cite{2024_agia_unpacking, duan2024aha} typically function as open-loop alarms or emergency stops~\cite{römerFailurePredictionRuntime2025}, lacking active correction capabilities.

\begin{wrapfigure}[14]{r}{0.65\textwidth}
    \centering
    \vspace{-15pt}
    \includegraphics[width=0.9\linewidth]{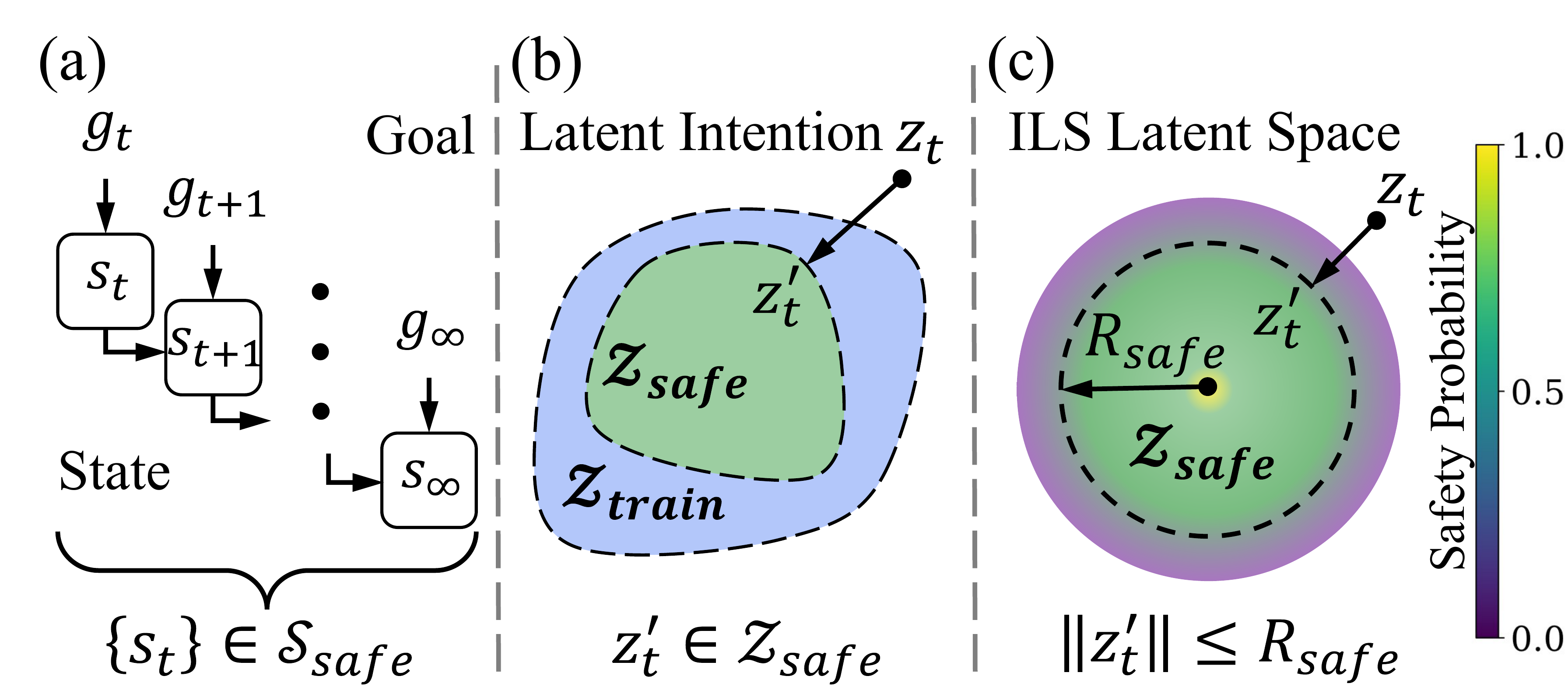}
    \vspace{-10pt}
    \caption{Evolution of the safety formulation. (a) The original problem involves infinite future horizons (Section~\ref{subsec:equivalence}). (b) We reduce this to a single-step latent inclusion problem, but face \emph{Boundary Ambiguity} between the training distribution and true safe set (Section~\ref{subsec:safety_training}). (c) ILS enforces a safety–radius isomorphism, yielding a spherical boundary with $\mathcal{O}(1)$ verification and correction via norm truncation (Section~\ref{subsec:isomorphism}).}
    \label{fig:method_progression}
\end{wrapfigure}

Recent inference-time steering methods~\cite{sunLatentPolicyBarrier2025, janner2022planning, ajay2022conditional, zhou2024diffusion, sun2024force, qi2025strengthening, nakamura2025generalizing, nakamura2025train} actively utilize gradient guidance or predictive models to enforce safety constraints or mitigate covariate shift~\cite{ross2011reduction}, but their reliance on iterative sampling introduces prohibitive latency for high-frequency whole-body control ($>50$~Hz)~\cite{luManiCMRealtime3D2025, weiClosedloopDiffusionControl2024}. In contrast, our $\mathcal{O}(1)$ projection enables seamless, high-frequency ``best-effort'' steering without iterative latency or disruptive emergency stops.

%% file: sections/03_method.tex
\section{Method}
\label{sec:method}

\begin{figure*}[!t]
    \centering
    \includegraphics[width=1.0\textwidth]{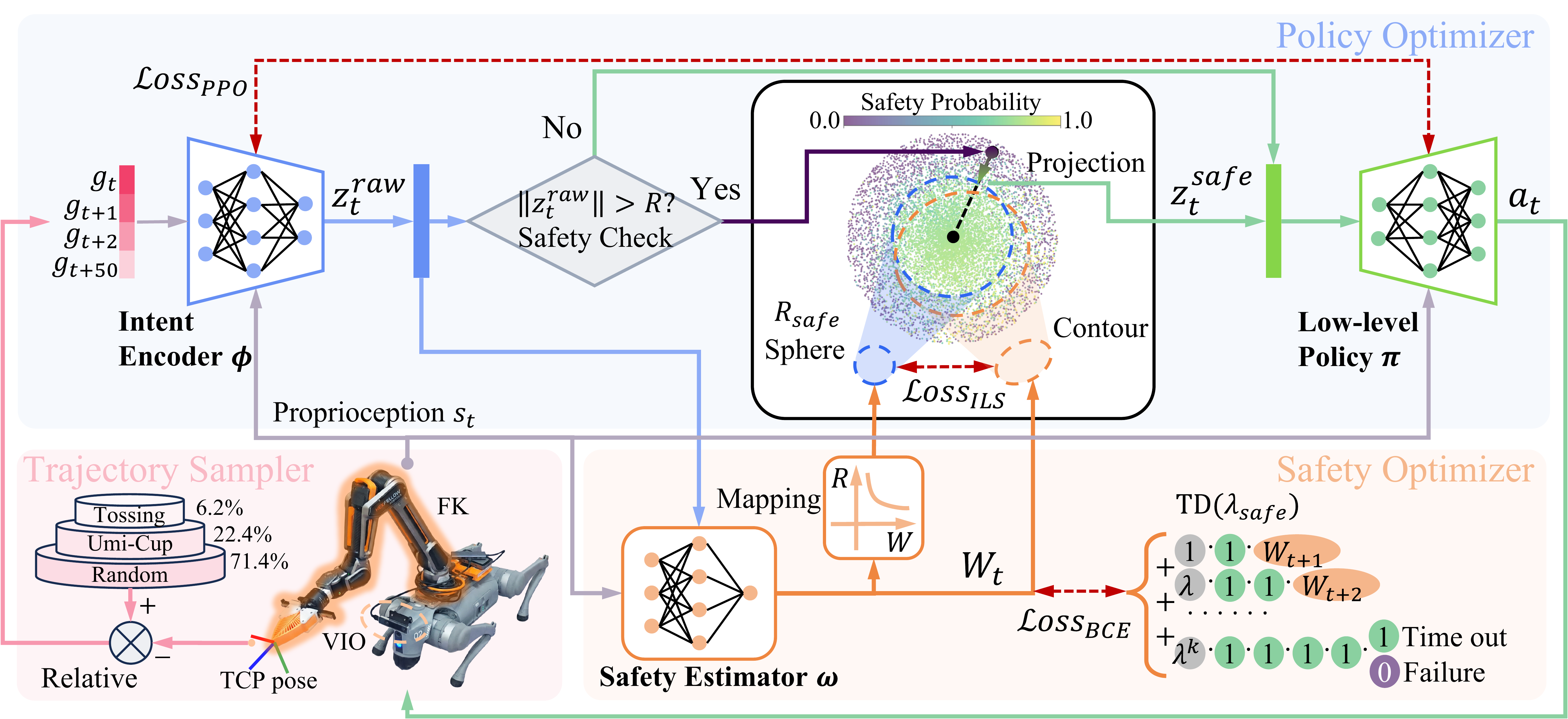}
    \vspace{-15pt}
    \caption{Overview of the pipeline. Target trajectories relative to the current Tool Center Point (TCP) frame are encoded into a raw latent intention $z_t^{raw}$ by an Intent Encoder $\phi$ for execution by the Low-level Policy $\pi$. A Safety Estimator $\omega$ is concurrently trained via TD targets to assess safety. To streamline inference, Isomorphic Latent Space (ILS) aligns safety contours to be spherical with safety decreasing radially. This permits $\mathcal{O}(1)$ safety enforcement without the estimator by simply truncating latent vectors that exceed the safety radius.}
    \label{fig:pipeline}
    \vspace{-10pt}
\end{figure*}

\subsection{Overview}
\label{subsec:method_overview}

We model the whole-body control task as a Markov Decision Process (MDP) defined by the tuple $(\mathcal{S}, \mathcal{A}, \mathcal{P}, \mathcal{G})$, where $\mathcal{S}, \mathcal{A}, \mathcal{P}$ denote the spaces of robot state, action, and environment dynamics, and $\mathcal{G}$ encapsulates task goals. We designate $\mathcal{S}_{safe} \subset \mathcal{S}$ as physically safe states constrained by limits such as torque saturation and contact forces. The objective is to ensure $\forall t \ge 0,\ s_t \in \mathcal{S}_{safe}$ for any initial state $s_0$ and arbitrary goal sequence $\mathbf{g} = \{g_t\}_{t=0}^\infty$, under policy $a_t \sim \pi(\cdot|s_t, g_t)$ and dynamics $s_{t+1} \sim \mathcal{P}(\cdot|s_t, a_t)$.

However, the policy input $(s_t, g_t)$---whether due to VIO estimation errors or aggressive user commands---may exceed the policy's competence, driving the system toward failure states $s \notin \mathcal{S}_{safe}$. Enforcing safety under these conditions presents three challenges:

\begin{itemize}
    \item \textbf{Temporal-Spatial Complexity (Fig.~\ref{fig:method_progression}a):} Safety is inherently coupled with the infinite-dimensional sequence $\mathbf{g}_{t:\infty} \in \mathcal{G}^\infty$. A command $g_t$ is safe only if there exists a future trajectory strictly within $\mathcal{S}_{safe}$. Since commands are end-effector \emph{trajectories} rather than simple setpoints, this feasible subspace is topologically fragmented and extremely sparse, making direct $\mathcal{O}(1)$ constraint in the original command space computationally intractable (see Appendix~\ref{app:necessity}).

    \item \textbf{Boundary Ambiguity (Fig.~\ref{fig:method_progression}b):} Let $\mathcal{D}_{safe} \subset \mathcal{S} \times \mathcal{G}$ denote the true set of safe state-goal pairs. Since our training distribution $\mathcal{D}_{train} \neq \mathcal{D}_{safe}$ (training data may contain unmastered failures), standard OOD detection relying on $(s_t, g_t) \in \mathcal{D}_{train}$ is insufficient for robustness.

    \item \textbf{Seamless Enforcement (Fig.~\ref{fig:method_progression}c):} Abrupt emergency stops cause instability and interrupt tasks, while post-fall recovery risks hardware damage. This requires finding the semantically closest safe command online to maintain stability. We seek an efficient projection $\Phi: \mathcal{G} \to \mathcal{G}$ mapping $g_t$ to $g'_t \in \mathcal{D}_{safe}$ that minimizes $\|g_t - g'_t\|_{semantic}$, enabling the closest safe command to be found online.
\end{itemize}

To address these challenges, we propose the \emph{Competence Manifold Projection} (CMP) framework. As illustrated in Fig.~\ref{fig:pipeline}, our architecture comprises an Intent Encoder $\phi$, a Low-level Policy $\pi$, and a Safety Estimator $\omega$. We train a unified policy for diverse trajectory tracking with a single set of network weights. Following \citet{Huy2024UMIonLegs}, we omit observation history and simplify notation to single-frame states.

\subsection{Frame-Wise Safety Scheme}
\label{subsec:equivalence}

This section addresses Challenge 1 by resolving the coupled temporal and spatial complexities of safety through a hierarchical latent framework.

To handle spatially fragmented safe regions in the original high-dimensional space, we map diverse target trajectories into a compact latent space, drawing inspiration from CVAEs~\cite{NIPS2015_8d55a249}. The Intent Encoder $\phi$ (MLP, hidden layers [256, 256]) encodes task goal $g_t$ conditioned on $s_t$ into a latent distribution $\mathcal{N}(\mu_\phi, \Sigma_\phi)$, from which latent command $z_t$ is sampled. The Low-level Policy $\pi$ (MLP, hidden layers [256, 256, 256]) then decides whole-body action $a_t$ from $z_t$ conditioned on $s_t$. Both $\pi$ and $\phi$ are jointly optimized via PPO~\cite{schulmanProximalPolicyOptimization2017a}. Crucially, both modules take $s_t$ as input, so the latent space purely encodes variation in future trajectory intent relative to the current state, allowing the distribution of safe trajectories to form a continuous, regular manifold in $\mathcal{Z}$.

This structure also enables eliminating the temporal dependency of safety assurance. By defining a suitable safety metric, we inherently reduce the intractable infinite-horizon safety problem into a verifiable single-step spatial inclusion condition. For any given $\pi(\cdot|s_t, z_t)$, we define the \emph{Maximum Probability of Perpetual Safety} $W^\pi(s_t, z_t)$: the probability of remaining in $\mathcal{S}_{safe}$ indefinitely when committing to $z_t$ now and selecting all future latent commands optimally. Formally,
\begin{equation}
    W^\pi(s_t, z_t) \triangleq \max_{\mathbf{z}_{t+1:\infty}} \mathbb{P} \left( \forall k \ge t, s_k \in \mathcal{S}_{safe} \;\middle|\; s_t, z_t, \mathbf{z}_{t+1:\infty}, \pi \right).
    \label{eq:safety_def}
\end{equation}
By the Markov property, this decomposes into the Bellman recursion:
\begin{equation}
    W^\pi(s_t, z_t) = \mathbb{I}_{\mathcal{S}_{safe}}(s_t) \cdot \mathbb{E}_{s_{t+1}} \left[ \max_{z' \in \mathcal{Z}} W^\pi(s_{t+1}, z') \right],
    \label{eq:safety_bellman}
\end{equation}
where the expectation is over $s_{t+1} \sim \mathcal{P}(\cdot|s_t, \pi(s_t, z_t))$.

Based on $W^\pi$, we define the \emph{Competence Manifold} $\mathcal{C}_\delta^\pi(s) \triangleq \{ z \in \mathcal{Z} \mid W^\pi(s, z) \ge 1-\delta \}$. The requirement that there exists at least one future sequence ensuring continuous safety with probability $\ge 1-\delta$ is exactly $W^\pi(s_t, z_t) \ge 1-\delta$, which directly implies the single-step inclusion $z_t \in \mathcal{C}_\delta^\pi(s_t)$. Thus, the complex infinite-horizon safety requirement reduces exactly to a verifiable single-step membership test---without quasi-static assumptions---making it well-suited for dynamic legged robots.

\subsection{Lower-Bounded Safety Estimator}
\label{subsec:safety_training}

While the scheme above connects safety to the Competence Manifold, its true boundary remains ambiguous since training data is not guaranteed to be perfectly safe. To identify mastered intentions, we train a Safety Estimator $\omega$ (MLP, hidden layers [256, 256]) to approximate $W^\pi(s_t, z_t)$, serving as a verifiable criterion beyond simple distribution matching.

Directly computing the Bellman update (Eq.~\eqref{eq:safety_bellman}) faces two hurdles: unknown transition dynamics and intractable maximization over continuous $\mathcal{Z}$. We address these through a conservative approximation strategy. First, we substitute $\max_{z'} W^\pi(s_{t+1}, z') \approx \widehat{W}_\omega(s_{t+1}, \mathbf{0})$, relying on the property that peak safety probabilistically aligns with the latent origin (see Appendix~\ref{app:proof_lower_bound}). Second, we employ TD($\lambda_{safe}$)~\cite{sutton2018reinforcement} to balance estimation bias and signal propagation: $n$-step expansion accelerates failure signal backpropagation but constitutes a lower bound that loosens with larger $n$, and TD($\lambda$) balances these multi-step returns for conservative estimation (proof in Appendix~\ref{app:proof_lower_bound}).

In practice, the training target $W_{target,t}$ is computed recursively backward through the trajectory:
\begin{equation}
    W_{target, t} = \begin{cases}
        0, & \text{if failure} \\
        1, & \text{if timeout} \\
        (1 - \lambda_{safe})\, \widehat{W}_\omega(s_{t+1}, \mathbf{0}) + \lambda_{safe}\, W_{target, t+1}, & \text{otherwise}
    \end{cases}
\end{equation}
with $\lambda_{safe}=0.8$. The estimator is then optimized via:
\begin{equation}
    \mathcal{L}_{BCE} = \text{BCE}\left( \widehat{W}_\omega(s_t, z_t),\ W_{target, t} \right).
\end{equation}

\subsection{Isomorphic Latent Space}
\label{subsec:isomorphism}

With the safety metric available, the remaining challenge is seamless enforcement under strict real-time constraints. Since online optimization of $W^\pi$ is computationally prohibitive, we propose \emph{Isomorphic Latent Space} (ILS), which reshapes the latent space to align safety levels with geometry, reducing safety assurance to efficient $\mathcal{O}(1)$ operations. Standard latent spaces use static KL divergence to cluster samples near the origin, whereas our objective demands safety decrease monotonically with radial distance, forming a spherical competence boundary. We exploit the concentration of high-dimensional Gaussian mass on spherical shells and dynamically modulate the KL prior, distributing intentions onto radial shells by safety probability to align geometric structure with safety while preserving semantic topology.

To map safety probability to geometry, we define a variance $R_t$ decreasing monotonically with $W_{target,t}$ via a cubic inverse formulation:
\begin{equation}
    R_t = \min \left( R_{\max}, \max \left( R_{\min}, \frac{1}{(W_{\text{target}, t} + \epsilon)^3} \right) \right),
\end{equation}
where $\epsilon = 1 - \bar{W}_{\text{batch}}$ centers the denominator around 1 for stable gradient scale, and min-max clipping prevents KL divergence explosion at extreme values. The Intent Encoder $\phi$ then minimizes:
\begin{equation}
    \mathcal{L}_{ILS} = D_{KL}\left( \mathcal{N}(\mu_\phi, \Sigma_\phi) \parallel \mathcal{N}(0, R_t^2 I) \right).
\end{equation}
Since $z \sim \mathcal{N}(0, R^2 I)$ concentrates around the shell $\|z\|_2 \approx \sqrt{d} R$, minimizing $\mathcal{L}_{ILS}$ organizes the latent space such that $\|z\| \propto R$: lower safety probabilities map to larger latent norms, reshaping irregular competence boundaries into regular hyperspheres.

At runtime, safety verification $z \in \mathcal{C}_\delta^\pi$ simplifies to the norm check $\|z_t^{raw}\| \le R_{safe}$, where $R_{safe}$ corresponds to the desired confidence $1-\delta$. When an OOD command yields a latent vector exceeding the competence boundary, we apply \emph{Competence Manifold Projection} (CMP):
\begin{equation}
    z_t^{safe} = z_t^{raw} \cdot \min\left(1, \frac{R_{safe}}{\|z_t^{raw}\|_2}\right).
\end{equation}
This bounds $\|z_t^{safe}\|_2 \le R_{safe}$, which by our frame-wise equivalence (Section~\ref{subsec:equivalence}) ensures a perpetually safe future trajectory ($P \ge 1-\delta$). Due to KL-enforced latent continuity, the projection approximates the \emph{closest feasible intent} rather than arbitrarily resetting the system, enabling ``best-effort'' tracking at the limit of the robot's capabilities. Visualizations of ILS are provided in Appendix~\ref{app:ils_visualization}.

%% file: sections/04_experiments.tex
\section{Experiments}

We evaluate CMP in simulation and on physical hardware using a Unitree Go2 quadruped equipped with a Hexfellow Saber robotic arm. Simulations are conducted in Isaac Gym~\cite{makoviychukIsaacGymHigh2021}. Baselines include UMI-on-Legs~\cite{Huy2024UMIonLegs}, Latent Shielding~\cite{nakamura2025generalizing}, and Neural CBF~\cite{nakamura2025train}; we exclude Safe RL and MPC: safe RL has no design for OOD inputs, and MPC handles OOD in principle but needs an accurate model and costly compute (Appendix~\ref{app:scope_composability}). We report Position Error ($e_p$), Orientation Error ($e_r$), and Survival Rate (SR), with episodes terminated upon physically unsafe conditions (see Appendix~\ref{app:termination_conditions}).

\subsection{Simulation Results}
\label{subsec:sim_overall}

We evaluate on 7,000 trajectories (13 seconds each) across three scenarios, all methods trained on the same dataset (see Appendix~\ref{app:dataset_generation}):
\begin{itemize}
    \item \textbf{In-Distribution (ID):} Targets mostly within frontal yaw $[-60^\circ, 60^\circ]$, comprising representative task trajectories with augmentation and random generation.
    \item \textbf{OOD-Geometry:} Target yaw $\in [120^\circ, 240^\circ]$ (rear), requiring geometric adaptation.
    \item \textbf{OOD-Sensor:} ID trajectories corrupted by sensor noise to simulate VIO failure.
\end{itemize}

Table~\ref{tab:sim_main_results} summarizes results. UMI-on-Legs achieves good ID tracking but fails catastrophically under OOD conditions. Neural CBF and Latent Shielding offer only marginal SR improvements: Neural CBF's Lie derivative conditions are frequently violated by complex legged dynamics, while Latent Shielding's hard thresholds abruptly interrupt tasks, creating a severe trade-off between tracking performance and survivability. CMP matches ID tracking precision while substantially boosting survival rates. At $R_{safe}=2.0$, CMP achieves a 10-fold SR improvement in OOD-Geometry (46.9\% vs.\ 4.7\%) and a nearly 6-fold increase under OOD-Sensor noise over UMI-on-Legs. Unlike abrupt shielding methods, CMP smoothly projects actions back to competence boundaries, preserving intent semantics while ensuring survival.

\begin{table*}[t]
    \centering
        \caption{Performance Comparison across ID and OOD Scenarios in Simulation.}
        \label{tab:sim_main_results}
        \resizebox{\linewidth}{!}{
        \begin{threeparttable}
            \begin{tabular}{ccccccccccc}
                \toprule
                \multirow{2}{*}{\textbf{Method}} & \multirow{2}{*}{$R_{safe}$} & \multicolumn{3}{c}{\textbf{In-Distribution (ID)}} & \multicolumn{3}{c}{\textbf{OOD-Geometry}} & \multicolumn{3}{c}{\textbf{OOD-Sensor}} \\
                \cmidrule(lr){3-5} \cmidrule(lr){6-8} \cmidrule(lr){9-11}
                & & SR (\%) $\uparrow$ & $e_p$ (cm) $\downarrow$ & $e_r$ (rad) $\downarrow$ & SR (\%) $\uparrow$ & $e_p$ (cm) $\downarrow$ & $e_r$ (rad) $\downarrow$ & SR (\%) $\uparrow$ & $e_p$ (cm) $\downarrow$ & $e_r$ (rad) $\downarrow$ \\
                \midrule
                UMI-on-Legs & - & 85.3 & 4.9 & 0.105 & 4.7 & - & - & 6.9 & - & - \\
                \midrule
                Latent Shielding & - & 90.9 & \textbf{4.4} & \textbf{0.096} & 5.6 & - & - & 17.7 & - & - \\
                \midrule
                Neural CBF & - & \textbf{95.9} & 5.4 & 0.123 & 7.1 & - & - & 19.8 & - & - \\
                \midrule
                \multirow{6}{*}{CMP (Ours)} & 1.0 & 84.2 & 15.6 & 0.291 & 82.5 & 102.2 & 1.935 & 56.7 & 27.9 & 0.687 \\
                & 1.5 & 95.2 & 6.0 & 0.139 & 62.4 & 90.8 & 1.767 & 49.7 & 15.9 & 0.475 \\
                & 2.0$^*$ & \textbf{94.7} & \textbf{4.5} & 0.106 & \textbf{46.9} & \textbf{79.0} & \textbf{1.661} & \textbf{40.3} & \textbf{11.8} & \textbf{0.367} \\
                & 2.5 & 91.3 & 4.1 & 0.097 & 34.1 & 69.4 & 1.612 & 34.5 & 9.9 & 0.308 \\
                & 3.0 & 94.7 & 3.8 & 0.091 & 26.3 & - & - & 29.3 & - & - \\
                & 100 & 94.9 & 3.5 & 0.086 & 5.1 & - & - & 16.1 & - & - \\
                \bottomrule
            \end{tabular}
            \begin{tablenotes}
                \footnotesize
                \item \emph{Note:} \textbf{Bold} indicates the best results among the baselines and the selected CMP configuration (denoted as $^*$), including ties within the omitted standard deviation margin ($<0.2$ cm or $0.0002$ rad). To avoid survivorship bias, tracking metrics ($e_p$ and $e_r$) are omitted (-) for SR $< 30\%$.
            \end{tablenotes}
        \end{threeparttable}
        }
        \vspace{-10pt}
\end{table*}

\subsection{Ablation Studies}
\label{subsec:ablation}

\begin{wraptable}[5]{r}{0.45\textwidth} %
    \centering
    \vspace{-22pt}
    \caption{Ablation Study Configurations.}
    \label{tab:ablation_setup}
    \resizebox{\linewidth}{!}{
    \begin{tabular}{lccc}
        \toprule
        \textbf{Method} & \textbf{Frame-Wise} & \textbf{Safety Estimator} & \textbf{ILS} \\
        \midrule
        UMI-on-Legs & $\times$ & $\times$ & $\times$ \\
        CVAE & \checkmark & $\times$ & $\times$ \\
        SCVAE & \checkmark & \checkmark & $\times$ \\
        CMP (Ours) & \checkmark & \checkmark & \checkmark \\
        \bottomrule
    \end{tabular}
    }
\end{wraptable}

We ablate three components (Table~\ref{tab:ablation_setup}): 
\textbf{CVAE} implements the Frame-Wise Safety Scheme (Section~\ref{subsec:equivalence}), truncating latent commands $z_t$ exceeding $R_{safe}$ during inference; 
\textbf{SCVAE} further adds the Safety Estimator (Section~\ref{subsec:safety_training}), applying KL divergence loss only to samples where $W(s_t, z_t)$ exceeds the batch average $\bar{W}_{\text{batch}}$---a direct filtering strategy without geometric shaping; 
\textbf{CMP} additionally employs Isomorphic Latent Space (ILS, Section~\ref{subsec:isomorphism}) to align latent geometry with safety probability.

\begin{wrapfigure}[14]{r}{0.5\textwidth}
    \centering
    \vspace{-10pt}
    \includegraphics[width=0.97\linewidth]{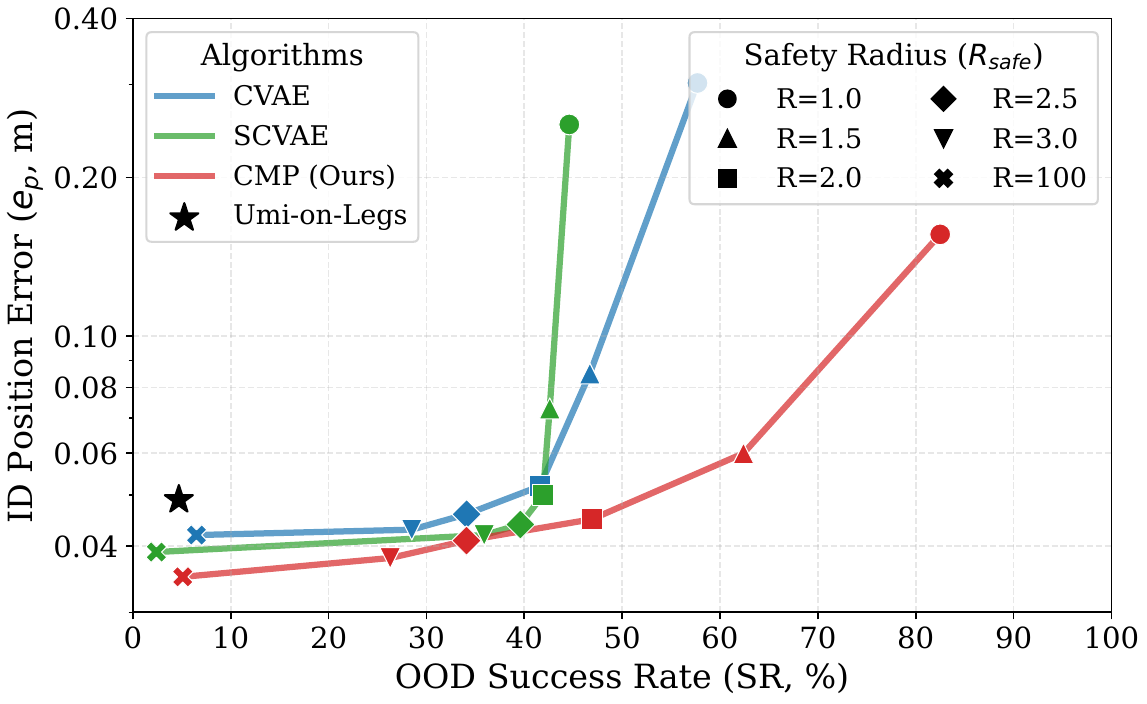}
    \vspace{-10pt}
    \caption{Accuracy--safety trade-off under varying $R_{safe}$. ID position error (log scale) is plotted against OOD-Geometry survival rate.}
    \label{fig:sim_error_safety}
\end{wrapfigure}

Sweeping $R_{safe}$ reveals the conservatism--agility trade-off (Fig.~\ref{fig:sim_error_safety}). SCVAE outperforms CVAE at larger radii but degrades below it at small radii, because its latent origin is neither density- nor safety-optimized: aggressive truncation yields codes that are neither accurate nor safe. CMP achieves a superior trade-off via ILS, which correlates safety with latent geometry and preserves richer functional behaviors at small radii where baselines rapidly degrade (further visualizations in Appendix~\ref{app:ils_visualization}). Qualitative and quantitative validation of the Safety Estimator $W(s, z)$ is provided in Appendix~\ref{app:safety_estimator_validation}.

\subsection{Real-World Experiments}
\label{sec:real_experiments}

Base pose is estimated via Visual-Inertial Odometry (VIO) from an onboard Intel RealSense T265, with pre-set target trajectories providing global target poses; the policy input is the computed relative target pose. OOD inputs arise from intrinsic sensor anomalies (e.g., VIO drift) and extrinsic infeasible user commands.

\begin{figure*}[t]
    \centering
    \vspace{5pt}
    \includegraphics[width=\textwidth]{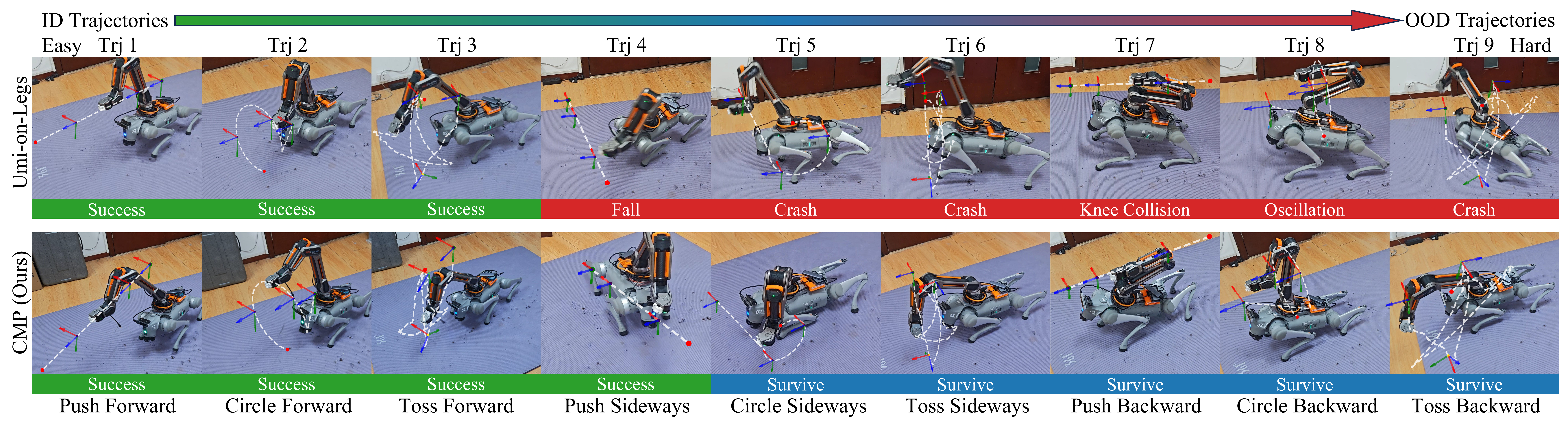}
    \vspace{-20pt}
    \caption{Robot behaviors under varying command difficulties (gripper removed to prevent damage). Color bars indicate outcomes: green for task success, blue for safe survival with reduced accuracy, and red for catastrophic failure. UMI-on-Legs fails catastrophically on OOD inputs; CMP generalizes to OOD tasks via ``best-effort'' behaviors.}
    \label{fig:real_exp_1}
    \vspace{-10pt}
\end{figure*}

Table~\ref{tab:real} evaluates 15 target trajectories on 4 methods across ID, Moderate OOD, and Extreme OOD categories---spanning spatial deviations (forward, sideways, backward) for tasks such as pushing and tossing, and dynamic variations (rapid jumping, fast cup manipulation). Fig.~\ref{fig:real_exp_1} visually compares two methods on 9 representative spatial tracking tasks.

Under \textbf{ID} conditions, CMP is the only method achieving 100\% SR (vs.\ $\sim$80\% for UMI-on-Legs and Neural CBF), with comparable tracking errors, confirming that safety projection does not hinder nominal precision. 
Under \textbf{Moderate OOD}---spatially or dynamically challenging tasks such as sideways tracking and diagonal jumping---UMI-on-Legs fails entirely (0\% SR) due to aggressive lateral actions destabilizing the base; Latent Shielding and Neural CBF achieve 33.3\% and 60.0\% SR; CMP reaches 93.3\% SR by projecting intentions onto the Competence Manifold, exhibiting ``best-effort'' behaviors (e.g., small safe turns for sideways pushing). 
Under \textbf{Extreme OOD}, tasks such as backward tracking and extreme side-jumping severely exceed the training distribution, and all baselines struggle (0.0\%, 20.0\%, 40.0\% SR), while CMP achieves 86.7\% SR by truncating unsafe command components to prioritize stability while preserving semantic intent.

\begin{table*}[t]
    \centering
    \caption{Performance Comparison on 15 Real-World Tasks across ID and OOD Scenarios.}
    \label{tab:real}
    \resizebox{\linewidth}{!}{
    \begin{threeparttable}
        \begin{tabular}{ccccccccccc}
            \toprule
            \multirow{2}{*}{\textbf{Method}} & \multicolumn{3}{c}{\textbf{ID} ($5$ tasks$\times$ $3$ trials)} & \multicolumn{3}{c}{\textbf{Moderate OOD} ($5$ tasks$\times$ $3$ trials)} & \multicolumn{3}{c}{\textbf{Extreme OOD} ($5$ tasks$\times$ $3$ trials)} & \multirow{2}{*}{Latency (ms) $\downarrow$} \\
            \cmidrule(lr){2-4} \cmidrule(lr){5-7} \cmidrule(lr){8-10}
            & SR (\%) $\uparrow$ & $e_p$ (cm) $\downarrow$ & $e_r$ (rad) $\downarrow$ & SR (\%) $\uparrow$ & $e_p$ (cm) $\downarrow$ & $e_r$ (rad) $\downarrow$ & SR (\%) $\uparrow$ & $e_p$ (cm) $\downarrow$ & $e_r$ (rad) $\downarrow$ & \\
            \midrule
            UMI-on-Legs & 80.0 & $\mathbf{4.9 \pm 1.6}$ & $\mathbf{0.07 \pm 0.02}$ & 0.0 & - & - & 0.0 & - & - & $\mathbf{2.97 \pm 0.15}$ \\
            \midrule
            Latent Shielding & 73.3 & $6.9 \pm 4.3$ & $0.09 \pm 0.03$ & 33.3 & $\mathbf{7.8 \pm 6.6}$ & $\mathbf{0.16 \pm 0.11}$ & 20.0 & - & - & $3.89 \pm 0.49$ \\
            \midrule
            Neural CBF & 80.0 & $\mathbf{4.8 \pm 1.9}$ & $0.09 \pm 0.03$ & 60.0 & $\mathbf{7.0 \pm 4.3}$ & $\mathbf{0.17 \pm 0.11}$ & 40.0 & $\mathbf{19.3 \pm 11.9}$ & $\mathbf{0.24 \pm 0.27}$ & $5.36 \pm 0.54$ \\
            \midrule
            CMP (Ours) & \textbf{100.0} & $\mathbf{5.1 \pm 1.8}$ & $0.09 \pm 0.03$ & \textbf{93.3} & $\mathbf{9.6 \pm 7.2}$ & $\mathbf{0.24 \pm 0.10}$ & \textbf{86.7} & $\mathbf{19.2 \pm 11.1}$ & $0.87 \pm 0.68$ & $\mathbf{2.99 \pm 0.14}$ \\
            \bottomrule
        \end{tabular}
        \begin{tablenotes}
            \footnotesize
            \item \emph{Note:} Best results \emph{and those within the error margin} are bolded. CMP utilizes $R_{safe}=2.0$ for all experiments. To avoid survivorship bias, tracking metrics ($e_p$ and $e_r$) are omitted (-) for SR $< 30\%$.
        \end{tablenotes}
    \end{threeparttable}
    }
    \vspace{-10pt}
\end{table*}

Beyond survival, Latent Shielding and Neural CBF incur latency overheads ($3.89$~ms and $5.36$~ms) detrimental to high-frequency whole-body control; CMP's implicit $\mathcal{O}(1)$ projection achieves $2.99$~ms, closely matching the unshielded baseline. Across 45 hardware trials, we observed 3 failures (see Section~\ref{sec:limitations}); robustness to sensor-induced OOD inputs from VIO drift is further demonstrated in Appendix~\ref{app:sensor_divergence}.

%% file: sections/05_conclusion.tex
\section{Conclusion}
\label{sec:conclusion}

We introduced Competence Manifold Projection (CMP) to secure whole-body controllers against OOD perturbations via three contributions: 
a Frame-Wise Safety Scheme reducing intractable infinite-horizon constraints to single-step latent inclusions, a Lower-Bounded Safety Estimator quantifying the maximum viability of arbitrary intents, and an Isomorphic Latent Space aligning this metric with latent geometry for $\mathcal{O}(1)$ projection. 
CMP achieves up to a \emph{10-fold improvement} in survival rates across typical OOD scenarios in simulation and real-world setups, incurring \emph{less than 10\%} in-distribution tracking degradation, while also exhibiting emergent ``best-effort'' behaviors that maximize task progress along the competence boundary.

%% file: sections/06_limitations.tex
\section{Limitations}
\label{sec:limitations}

Among the 45 hardware trials, 3 failures were observed: 
the nature of network learning, compounded by the theoretical lower bound (Appendix~\ref{app:proof_lower_bound}) and the empirical approximation (Appendix~\ref{app:safety_estimator_validation}), can cause the Safety Estimator to over-predict safety, producing oscillatory motions (1 occurrence); 
and while the spherical competence boundary holds statistically, it may fail to salvage execution in rare edge cases where projected commands remain unsafe (2 occurrences). 
Looking forward, achieving competence-aware control under dynamics shifts in high-dimensional humanoid systems remains an open challenge; CMP's architecture naturally extends toward this goal via environment conditioning. As an end-to-end learned policy, CMP brings safety closer to an absolute guarantee than ever, and further refinement of the ILS mechanism promises to narrow that gap.

%% file: sections/07_acknowledgments.tex
\acknowledgments{This work was supported in part by the National Key Research and Development Program of China under Grant 2022ZD0114903, in part by the National Natural Science Foundation of China under Grant 62336004, in part by Beijing Natural Science Foundation under Grant L247009, and in part by the Beijing Major Science and Technology Project under Contract no. Z251100008125019 and no. Z251100008125064.}

%% file: sections/08_appendix.tex
\appendix

\section{Necessity of Latent Space Safety}
\label{app:necessity}

Unlike simpler relative-to-base tracking methods, modern task-space Whole-Body Control paradigms (compatible with UMI~\cite{chi2024umi} and FastUMI~\cite{Zhaxizhuoma2025FastUMI}) receive full end-effector trajectories expressed in global or TCP coordinate systems. These commands typically consist of multiple keyframes spanning the full 6 Degrees of Freedom (e.g., 6-DoF $\times$ 4 keyframes = 24-dimensional space). This dramatic increase in command dimensionality makes naive solutions to OOD robustness unviable:

\begin{figure}[ht]
    \centering
    \includegraphics[width=0.9\linewidth]{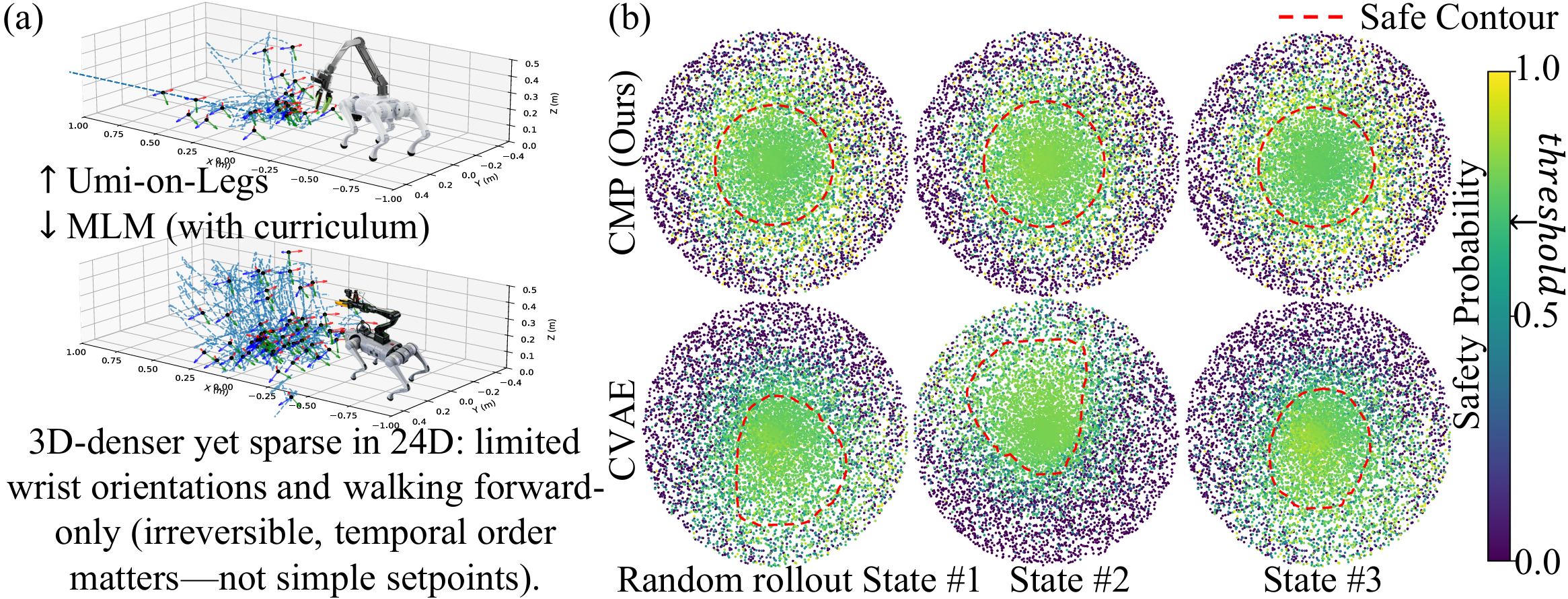}
    \caption{(a) Extreme sparsity of competent trajectories in the 24D command space even with curriculum learning. (b) Latent space visualization confirming our ILS mechanism effectively reshapes intents into a statistical spherical boundary, tightening the $\max$ approximation at the origin.}
    \label{fig:core_mechanisms}
\end{figure}

\begin{itemize}
    \item \textbf{Command-Space Constraints:} The feasible region in this 24D space exhibits extreme sparsity and fragmentation. A minor modification to a coordinate may cause the arm to hit a singularity, necessitating an entirely distinct system-level solution (such as a 180-degree base reorientation) to track properly. Furthermore, temporal order matters significantly, as trajectories are fundamentally distinct from simple setpoints. For instance, successfully learning to track a trajectory forward does not imply the ability to track it in reverse. Consequently, direct $\mathcal{O}(1)$ feasibility verification in the raw command space is intractable.
    \item \textbf{Curriculum Learning:} Unlike simple setpoint reaching, trajectory tracking requires expert trajectories in RL as reasonable references. As visualized in Fig.~\ref{fig:core_mechanisms}a, prior works such as MLM~\cite{liuMLMLearningMultitask2025} employ curriculum learning to acquire richer trajectories than UMI-on-Legs~\cite{Huy2024UMIonLegs}, but still fall far from full coverage of the command space. The mastered command distribution may seem dense in 3D space, but it remains extremely sparse in the 24D space.
\end{itemize}

\section{Visualization of Isomorphic Latent Space}
\label{app:ils_visualization}

To empirically verify the spatial organization governed by our Isomorphic Latent Space (ILS), we utilized PCA to visualize the latent space of 3 random states (Fig.~\ref{fig:core_mechanisms}b). 
As theoretically modelled in Section~\ref{subsec:isomorphism}, ILS establishes a dynamic inverse correlation such that low safety predictions induce expansive radii through a monotonically decreasing KL divergence boundary $R_t$. From the properties of high-dimensional Gaussians ($z \sim \mathcal{N}(0, R_t^2 I)$), probability mass concentrates heavily on spherical shells corresponding to their safety estimates, statistically aligning safety levels with geometry.

\section{Proof of Lower Bound Property}
\label{app:proof_lower_bound}

This section provides the theoretical proof from Section~\ref{subsec:safety_training}, verifying that the $n$-step expansion and the TD($\lambda_{safe}$) training target constitute a strict lower bound of the true safety probability.

\subsection{Definitions}

Let $\mathbb{I}(s)$ denote the safety indicator $\mathbb{I}_{\mathcal{S}_{safe}}(s)$.
We define the \emph{Optimal Safety Value} at state $s$ as the maximum safety probability achievable from $s$:
\begin{equation}
    V^*(s) \triangleq \max_{z \in \mathcal{Z}} W^\pi(s, z).
\end{equation}
The multiplicative Bellman equation (Eq.~\eqref{eq:safety_bellman}) can be rewritten using this notation as:
\begin{equation}
    W^\pi(s_t, z_t) = \mathbb{I}(s_t) \cdot \mathbb{E}_{s_{t+1}|s_t, z_t} \left[ V^*(s_{t+1}) \right].
    \label{eq:app_bellman}
\end{equation}

\subsection{\texorpdfstring{The $n$-step Expansion Operator}{The n-step Expansion Operator}}

The \emph{$n$-step probability expansion operator} $\mathcal{P}^{(n)}$ is defined as the expected safety calculated using the rollout trajectory $\tau \sim \mathcal{D}_{rollout}$ for the first $n$ steps and assuming optimal control thereafter:
\begin{equation}
    \mathcal{P}^{(n)} \triangleq \mathbb{E}_{\tau} \left[ \left( \prod_{i=0}^{n-1} \mathbb{I}(s_{t+i}) \right) V^*(s_{t+n}) \right].
\end{equation}
For $n=1$, this simplifies to $\mathcal{P}^{(1)} = W^\pi(s_t, z_t)$.

\subsection{Monotonicity Proof}

We now prove that the sequence is monotonically non-increasing, i.e., $\mathcal{P}^{(n)} \ge \mathcal{P}^{(n+1)}$.
First, we expand $V^*(s_{t+n})$ using the Bellman equation:
\begin{align}
    V^*(s_{t+n}) &= \max_{z'} \left( \mathbb{I}(s_{t+n}) \mathbb{E}_{s'| s_{t+n}, z'} \left[ V^*(s') \right] \right) \nonumber \\
                &\ge \mathbb{I}(s_{t+n}) \cdot \mathbb{E}_{s_{t+n+1} | s_{t+n}, z_{t+n}} \left[ V^*(s_{t+n+1}) \right].
    \label{eq:value_inequality}
\end{align}
The inequality holds because the maximum over all $z'$ is inherently greater than or equal to the expectation over the specific intention $z_{t+n}$ sampled from the rollout policy's encoder.

Substituting Eq.~\eqref{eq:value_inequality} back into the definition of $\mathcal{P}^{(n)}$:
\begin{align}
    \mathcal{P}^{(n)} &\ge \mathbb{E}_{\tau} \left[ \left( \prod_{i=0}^{n-1} \mathbb{I}(s_{t+i}) \right) \mathbb{I}(s_{t+n}) \mathbb{E}_{s_{t+n+1}} [ V^*(s_{t+n+1}) ] \right] \nonumber \\
    &= \mathbb{E}_{\tau} \left[ \left( \prod_{i=0}^{n} \mathbb{I}(s_{t+i}) \right) V^*(s_{t+n+1}) \right] \nonumber \\
    &= \mathcal{P}^{(n+1)}.
\end{align}
This establishes the chain $W^\pi(s_t, z_t) = \mathcal{P}^{(1)} \ge \mathcal{P}^{(2)} \ge \dots \ge \mathcal{P}^{(n)}$, confirming that the $n$-step rollout provides a lower bound estimation.

\subsection{\texorpdfstring{TD($\lambda_{safe}$) Target and Tightness}{TD(lambda\_safe) Target and Tightness}}

The training target $G_t^{\lambda_{safe}}$ is defined as the geometric weighted average of these $n$-step expansions:
\begin{equation}
    G_t^{\lambda_{safe}} = (1 - \lambda_{safe}) \sum_{n=1}^{\infty} \lambda_{safe}^{n-1} \mathcal{P}^{(n)}.
\end{equation}
Since $\{\mathcal{P}^{(n)}\}$ is a monotonically non-increasing sequence, the convex combination is upper-bounded by its first term:
\begin{equation}
    G_t^{\lambda_{safe}} \le \mathcal{P}^{(1)} = W^\pi(s_t, z_t).
\end{equation}
Thus, $G_t^{\lambda_{safe}}$ remains a strict lower bound of the true optimal safety.

The choice of $\lambda_{safe}$ involves a trade-off among four sources of error:

\begin{itemize}
    \item \textbf{Propagation Delay}: $\lambda_{safe}$ determines the weight of multi-step returns. As $\lambda_{safe} \to 0$, the target degenerates to single-step bootstrapping $\mathbb{I}(s_t) V^*(s_{t+1})$, leading to slow back-propagation of failure signals. Increasing $\lambda_{safe}$ enables direct signal propagation across time steps via eligibility traces.
    \item \textbf{Variance}: As $\lambda_{safe} \to 1$, the target relies on longer stochastic rollout trajectories. Due to the randomness in environment dynamics and policy execution, the variance of the cumulative probability estimate increases significantly.
    \item \textbf{Estimator Bias}: The term $V^*(s_{t+n})$ in $\mathcal{P}^{(n)}$ relies on the approximation by the Safety Estimator network $\omega$. A smaller $\lambda_{safe}$ causes earlier bootstrapping, making the target highly dependent on the potentially inaccurate $\omega$.
    \item \textbf{Lower Bound Gap}: As $\lambda_{safe}$ increases, the weight shifts towards $\mathcal{P}^{(\infty)}$. This implies that the estimated value transitions from the ``theoretical optimal safety'' to the ``on-policy safety,'' which constitutes a more conservative lower bound.
\end{itemize}

We select $\lambda_{safe} = 0.8$. This value strikes a balance between suppressing estimator bias and controlling sampling variance, while enabling rapid back-propagation of failure signals.

\subsection{Approximation of the Maximization}

The substitution of $\max_{z'} W^\pi(s_{t+1}, z')$ with the value at the latent origin $\widehat{W}_\omega(s_{t+1}, \mathbf{0})$ introduces an approximation. 
Theoretically, because $\max_{z'} W^\pi(s_{t+1}, z') \ge W^\pi(s_{t+1}, \mathbf{0})$, this substitution yields an even more conservative learning target $W_{\text{target},t} \le \max_{z'} W^\pi(s_{t+1}, z')$, reinforcing the theoretical lower-bound sequence proved above. 

Practically, this bound is tight (Fig.~\ref{fig:core_mechanisms}b showing the safety at origin is very close to peak). Initially, the KL divergence inherently tends to encode frequently executed actions near the origin. As the RL policy improves, these frequently chosen actions naturally correspond to relatively safe and competent behaviors. This serves as an initial training signal for the Safety Estimator. Then a virtuous training circle emerges: the initially trained estimator guides the Isomorphic Latent Space (ILS) to further concentrate safer intentions to the center. This in turn tightens the assumption $\max_{z'} W^\pi(s_{t+1}, z') \approx \widehat{W}_\omega(s_{t+1}, \mathbf{0})$, which subsequently provides more accurate targets for the estimator, leading to progressively better training of both the estimator and the latent space.

\section{Validation of Safety Estimator}
\label{app:safety_estimator_validation}

\subsection{Qualitative Validation via Rollout Analysis}

\begin{figure}[ht]
    \centering
    \includegraphics[width=0.6\linewidth]{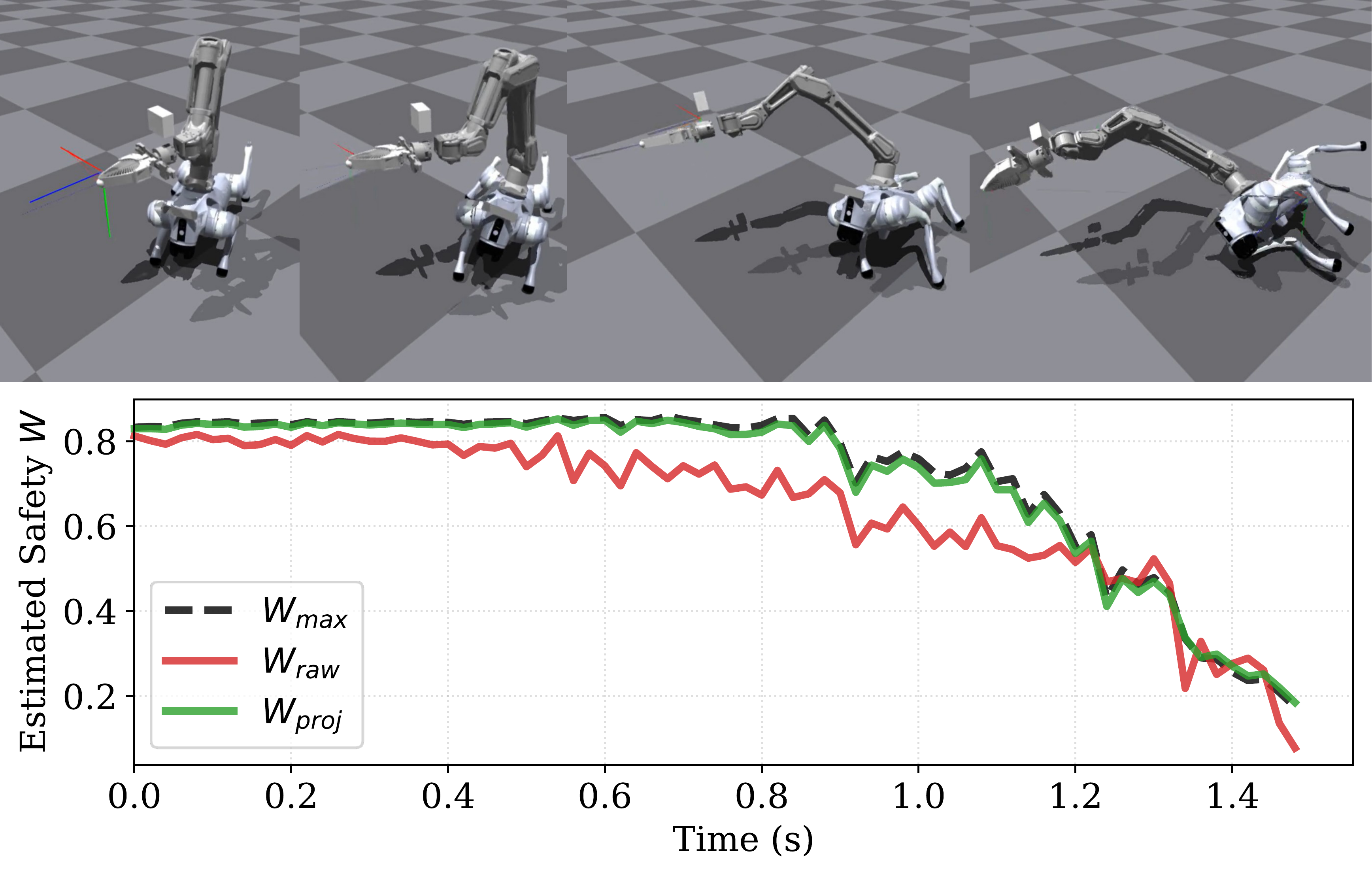}
    \caption{Validation of the Safety Estimator. Top: Snapshots of the robot executing a raw OOD sideways push command without latent projection. Bottom: Time-series of Safety metric $W$ of the safest intention, raw input intention and the projected intention.}
    \label{fig:sim_safety}
\end{figure}

We use a rollout trajectory to qualitatively validate the effectiveness of the Safety Estimator $W(s,z)$. We execute the policy using raw latent codes $z_t^{raw}$ without safety projection, allowing us to observe the safety metric's response to dangerous behaviors. Three phases are observed in Fig.~\ref{fig:sim_safety}:
\begin{enumerate}
    \item \textbf{Normal State (e.g., t=0.2s):} The robot tracks a feasible target. Safety metrics $W_{max}=W(s,\mathbf{0})$, $W_{proj}=W(s, z_t^{safe})$, and $W_{raw}=W(s, z_t^{raw})$ are all high. Since $z_t^{raw}$ is safe, $W_{max} > W_{raw} \approx W_{proj}$, validating the safety assessment.
    \item \textbf{OOD Target (e.g., t=1.0s):} The target becomes OOD. We observe $W_{max} > W_{proj} > W_{raw}$, indicating the estimator correctly penalizes the risky $z_t^{raw}$, while projection yields a safer $z_t^{safe}$. This validates the sensitivity of $W$ to $z$.
    \item \textbf{Near Fall (e.g., t=1.4s):} Unsafe actions drive the robot to a near-fall state. All $W$ values drop significantly, confirming $W(s,z)$ effectively captures state-dependent risks.
\end{enumerate}

\subsection{Quantitative Validation via Human Study}

Our Safety Estimator $W(s, z)$ intrinsically targets the metric formalized in Eq.~\eqref{eq:safety_def}: whether \emph{any} command exists that can salvage the state from failure. Specifically, computing this exact conditional maximum $\max W$ analytically is computationally intractable, making ground truth comparisons practically impossible. To validate the estimator, we instead correlate the network's predictions against human labels.

In Table~\ref{tab:human_study}, we randomly sampled 500 rollout states from OOD executions. For each state, 5 independent human evaluators provided a binary vote: ``Salvageable'' (could the robot recover if the best possible command sequence were given?) or ``Unsalvageable'' (is a fall inevitable regardless of future inputs?). The Safety Estimator scores $W_{\max} = \widehat{W}_\omega(s_{t+1}, \mathbf{0})$ were logged.

\begin{table}[ht]
\centering
\caption{Estimator Validation (500 random rollout states, 5 voters)}
\label{tab:human_study}
\resizebox{0.6\columnwidth}{!}{%
\begin{threeparttable}
\begin{tabular}{lccc}
\toprule
Human Label & $W_{\max} < 0.6$ & $0.6 \leq W_{\max} \leq 0.8$ & $W_{\max} > 0.8$ \\
\midrule
Salvageable  & \textbf{0.4\%}  & 9.2\% & \textbf{57.8\%} \\
Unsalvageable & \textbf{16.8\%} & 14.4\% & \textbf{1.4\%}  \\
\bottomrule
\end{tabular}%
\end{threeparttable}
}
\end{table}

When the estimator evaluates that a state has low probability to be salvageable ($W_{\max} < 0.6$), the conditional probability that a human perceives the situation as unsalvageable reaches $16.8\% / (16.8\% + 0.4\%) \approx 97.7\%$. Conversely, when the estimator implies high certainty in safety ($W_{\max} > 0.8$), the probability that humans confirm it is indeed salvageable is $57.8\% / (57.8\% + 1.4\%) \approx 97.6\%$. This strong correlation validates that our Safety Estimator effectively captures the underlying safety semantics as perceived by human evaluators, confirming its practical utility for real-time safety assessment and correction in OOD scenarios.

\section{Robustness to Sensor Divergence}
\label{app:sensor_divergence}

\begin{figure}[ht]
    \centering
    \includegraphics[width=0.6\linewidth]{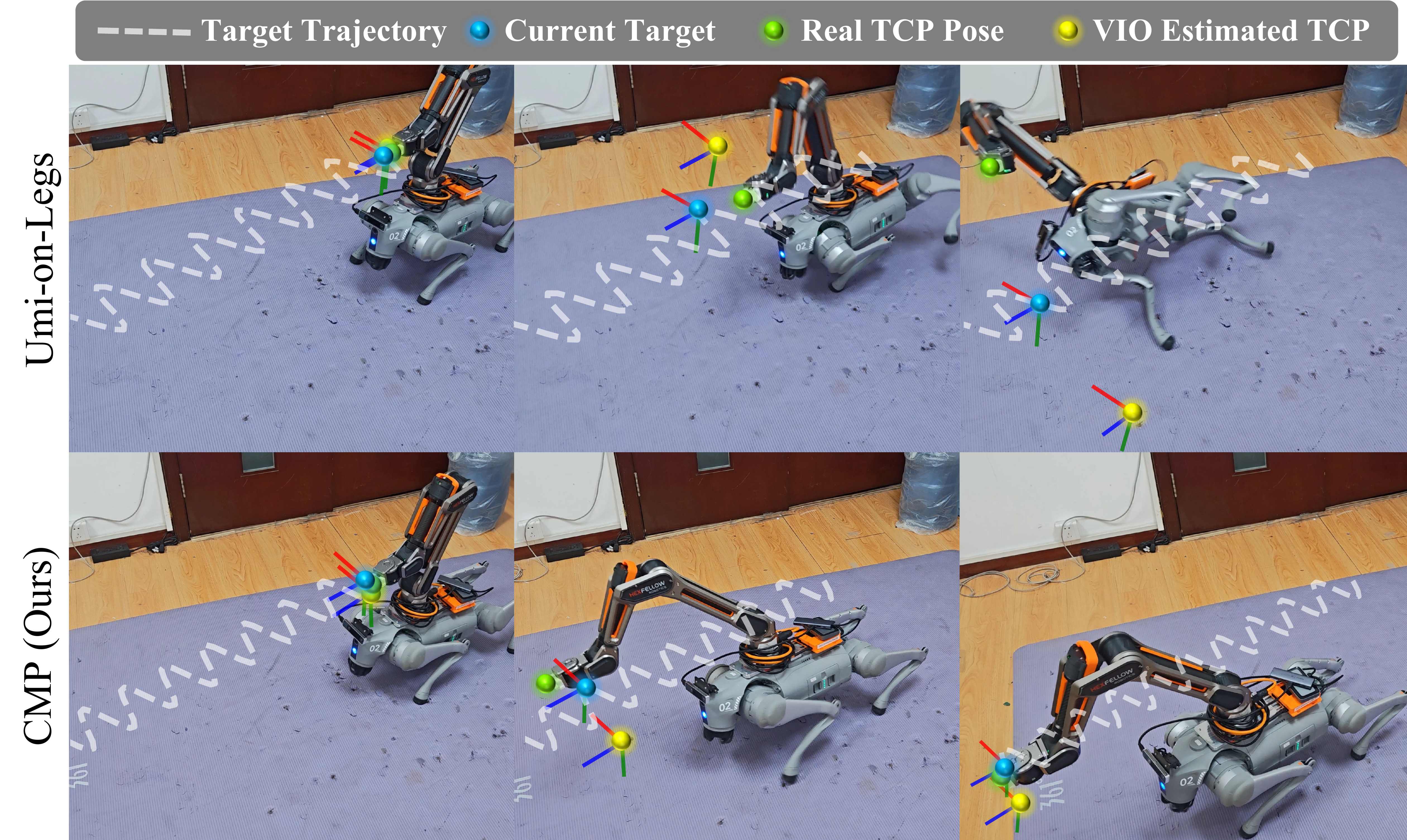}
    \caption{Defense against sensor-induced divergence. (Top) UMI-on-Legs~\cite{Huy2024UMIonLegs} enters a positive feedback loop: VIO error $\rightarrow$ Unexpected input $\rightarrow$ Aggressive motion $\rightarrow$ Larger VIO error $\rightarrow$ Crash. (Bottom) CMP prevents the unexpected inputs from triggering unsafe motion, effectively blocking the hazardous feedback loop and preserving stability.}
    \label{fig:real_exp_2}
\end{figure}

We investigate robustness against sensor-induced OOD inputs by commanding the robot to track a sinusoidal forward trajectory that necessitates rapid lateral body adjustments. Such rapid motion often causes the VIO odometry to jitter or drift significantly.

Fig.~\ref{fig:real_exp_2} illustrates the critical divergence mechanism triggered by the T265 sensor's bandwidth limitation:
(1) Rapid oscillation induces VIO drift, creating an erroneous, distorted relative goal $g_t$.
(2) For UMI-on-Legs~\cite{Huy2024UMIonLegs}, this OOD goal elicits aggressive corrective actions.
(3) These actions intensify body oscillation, forming a positive feedback loop that rapidly destabilizes the system (Fig.~\ref{fig:real_exp_2} Top).

In contrast, CMP (Fig.~\ref{fig:real_exp_2} Bottom) detects the low survival probability associated with the anomalous goal and projects the latent command to a safe region, dampening the response to sensor noise. This effectively blocks the dangerous feedback loop, preventing error amplification and maintaining stability.

This behavior is consistent with the OOD-Sensor simulation results in Table~\ref{tab:sim_main_results}: CMP achieves nearly a 6-fold improvement in survival rate over UMI-on-Legs under sensor noise injection, demonstrating that the competence boundary provides a principled defense against both extrinsic command OOD and intrinsic sensor anomalies.

\section{Dataset Generation Details}
\label{app:dataset_generation}

To ensure the diversity and robustness of the learned policy, we curate a composite dataset derived from both real-world human demonstrations and procedural generation. All trajectories are unified to a fixed duration of $T=2500$ steps at 200 Hz (12.5 seconds). We define the global coordinate system such that the robot base initially faces the $+x$ direction, with gravity acting along the $-z$ axis.

\subsection{In-Distribution (ID) Dataset}
The training dataset consists of 7,000 trajectories, constructed from two primary sources:

\paragraph{Augmented UMI Data}
We utilize the open-source dataset from UMI-on-Legs~\cite{Huy2024UMIonLegs}, specifically incorporating 1,090 ``cup in the wild'' trajectories and 101 ``tossing'' trajectories collected via the UMI~\cite{chi2024umi} data collection pipeline. To balance the distribution, the tossing subset is oversampled by a factor of three. We sample 2,000 trajectories from this collected pool and apply rigid body transformations to augment the workspace coverage:
\begin{itemize}
    \item \textbf{Centering \& Translation:} Trajectories are first centered at the origin, then translated along the x-axis by a random offset $\delta_x \sim \mathcal{U}(-0.2, 0.2)$\,m.
    \item \textbf{Rotation:} We apply a random yaw rotation $\theta \sim \mathcal{U}(-30, 30)^\circ$ around a pivot point defined at $(-0.3, 0, 0)$\,m relative to the robot base. This simulates variations in task orientation within the frontal workspace.
\end{itemize}

\paragraph{Procedural Random Pushes}
To enhance the policy's tracking capability across the full kinematic range, we generate 5,000 synthetic trajectories.
\begin{itemize}
    \item \textbf{Position:} The end-effector follows a random walk sequence. In the XY plane, each way-point is generated by moving a random distance $d \sim \mathcal{U}(0.1, 0.5)$\,m towards a direction $\phi \sim \mathcal{U}(-45, 45)^\circ$ relative to the forward ($+x$) axis, with a travel speed sampled from $\mathcal{U}(0.01, 0.4)$\,m/s. The Z-axis movements vary independently within $[0.02, 0.6]$\,m, with vertical speeds sampled from $\mathcal{U}(0.01, 0.2)$\,m/s.
    \item \textbf{Orientation:} Target orientations are generated by linearly interpolating between random Euler configurations ($\text{Roll} \in [-30, 30]^\circ$, $\text{Pitch} \in [15, 60]^\circ$, and $\text{Yaw} \in [-45, 45]^\circ$). The interpolation speed (angular velocity) is randomized for each segment, sampled from $\mathcal{U}(0.01, 1.0)$\,rad/s.
\end{itemize}

An independent ID test set is generated using the same protocol but with different random seeds.

\subsection{Out-of-Distribution (OOD) Datasets}
We evaluate generalization using two challenging variants, each containing 7,000 trajectories derived from the full ID dataset.

\paragraph{OOD-Geometry (Rear Workspace)}
This dataset evaluates the agent's competence in reaching targets completely outside the training distribution (specifically, behind the robot). We take the completed ID dataset as a base and re-apply the augmentation pipeline described above, sampling 7,000 times, but modifying the rotation parameter to sample $\theta \sim \mathcal{U}(179, 181)^\circ$. This effectively mirrors the entire distribution of frontal tasks to the rear of the robot, requiring significant whole-body reorientation.

\paragraph{OOD-Sensor (Drift \& Jumps)}
This dataset simulates severe state estimation failures such as VIO drift. We process every trajectory in the ID dataset by injecting discrete drift events at random intervals $\Delta t \in [1, 5]$\,s. At each event, a persistent bias is added to the remainder of the trajectory:
\begin{itemize}
    \item \textbf{Position Drift:} Additive Gaussian noise $\delta_p \sim \mathcal{N}(0, 0.2^2)$\,m.
    \item \textbf{Orientation Drift:} Multiplicative rotation noise derived from Euler angles $\delta_r \sim \mathcal{N}(0, 30^2)^\circ$.
\end{itemize}

\section{Termination Conditions}
\label{app:termination_conditions}
We largely follow the termination protocols established in UMI-on-Legs~\cite{Huy2024UMIonLegs} to define the safety boundary $\mathcal{S}_{safe}$. An episode is terminated immediately as a failure if any of the following physical safety constraints are violated specifically:

\begin{itemize}
    \item \textbf{Invalid Body Contacts (Falls):} A fall is detected if any risk-sensitive rigid body comes into contact with the environment (ground) with a force magnitude exceeding $1.0$\,N. The specific links triggering termination are:
    \begin{itemize}
        \item \textbf{Base \& Torso:} Base configuration links, Hip links, Thigh links, and the Head.
        \item \textbf{Manipulator Arm:} All arm segments, including the Base Arm Link and Links 1-6.
    \end{itemize}
    Note that the feet are explicitly permitted to contact the ground to support locomotion.
\end{itemize}

Other operational constraints, such as joint limits, torque saturation, and action rates, are configured as soft constraints. Violations of these limits result in negative reward penalties rather than episode termination.

\section{Implementation Details}
\label{app:implementation}

All algorithms are implemented in PyTorch and trained on a single NVIDIA RTX A6000 GPU. The total training duration is approximately 2.5 hours for 2,000 iterations.

\subsection{Network Architectures}
We implement all network modules using Multi-Layer Perceptrons (MLPs). The detailed architectural configurations are summarized in Table~\ref{tab:network_arch}.

\begin{table}[ht]
    \centering
    \caption{Network Architectures}
    \label{tab:network_arch}
    \resizebox{0.6\linewidth}{!}{
    \begin{threeparttable}
            \begin{tabular}{lcccc}
                \toprule
                \textbf{Module} & \textbf{Input} & \textbf{Hidden} & \textbf{Output} & \textbf{Activation} \\
                \midrule
                Encoder $\phi$ & $s_t, g_t$ & $[256, 256]$ & 6 & ELU \\
                Policy $\pi$ & $s_t, z_t$ & $[256, 256, 256]$ & 12+6 & ELU \\
                Safety $\omega$ & $s_t, z_t$ & $[256, 256]$ & 1 & ELU+Sig\tnote{*} \\
                \bottomrule
            \end{tabular}
        \begin{tablenotes}
            \footnotesize
            \item[*] ELU for hidden layers; Sigmoid for the output layer.
        \end{tablenotes}
    \end{threeparttable}
    }
\end{table}

\subsection{Training Hyperparameters}
The training process involves separate optimizers for the policy and the Safety Estimator. We utilize a reward formulation consistent with UMI-on-Legs~\cite{Huy2024UMIonLegs} for the PPO surrogate loss. A detailed summary of hyperparameters is provided in Table~\ref{tab:hyperparameters}.

\begin{table}[ht]
    \centering
    \caption{Training Hyperparameters}
    \label{tab:hyperparameters}
    \resizebox{0.6\linewidth}{!}{
    \begin{threeparttable}
            \begin{tabular}{lcc}
                \toprule
                \textbf{Parameter} & \textbf{Policy Opt. ($\pi, \phi$)} & \textbf{Safety Opt. ($\omega$)} \\
                \midrule
                Optimizer & Adam & Adam \\
                Num. Environments & \multicolumn{2}{c}{4,096} \\
                Total Iterations & \multicolumn{2}{c}{2,000} \\
                PPO Epochs & \multicolumn{2}{c}{32} \\
                Mini-batches & \multicolumn{2}{c}{4} \\
                Learning Rate (LR) & Adaptive $1\text{e-}3$ & Fixed $1\text{e-}3$ \\
                Discount ($\gamma$) & 0.9 & - \\
                GAE ($\lambda$) & 0.95 & - \\
                Clip Range ($\epsilon$) & 0.2 & - \\
                KL Coef. ($\beta_{KL}$) & $0 \xrightarrow{1k} 1\text{e-}3$ & - \\
                Safety Disc. ($\lambda_{safe}$) & - & 0.8 \\
                Loss Function & $\mathcal{L}_{PPO} + \beta_{KL}\mathcal{L}_{ILS}$ & $\mathcal{L}_{BCE}$ \\
                \bottomrule
            \end{tabular}
    \end{threeparttable}
    }
\end{table}

\section{Scope and Composability}
\label{app:scope_composability}

We clarify the scope of our robustness claim and its composability with complementary approaches. The hazard set $\mathcal{S}_{safe}$ is not fixed: it enters only via the termination flag---torque limits, self-collision, and longer-horizon contact need no method change. We address robustness to the \emph{input command}---a source prior work has not targeted---not to dynamics shift; the two are complementary and fuse readily: CMP accepts terrain conditioning or randomized disturbance like any RL method. Payload, terrain, or attachment sweeps grade that method, not our filter, hence their absence from our evaluation. Safe RL~\cite{OpenAI2019SafeRL, brunke2022safe} complements us likewise: it makes the policy safer \emph{within} its distribution, but has no design for an OOD input, on which the network still acts unpredictably and potentially unsafely. MPC~\cite{Bellicoso2019ALMA, Sleiman2023Versatile} indeed handles OOD in principle, but needs an accurate model and costly compute; CMP avoids both while retaining the composability noted above. CMP therefore specifically robustifies \emph{model-free}, end-to-end RL; adopting an explicit model is an orthogonal direction, outside our scope. We also note that guarantee-like wording should be read statistically: CMP \emph{complements, rather than replaces}, external safety monitors.

\section{Spherical Boundary and Origin Verification}
\label{app:sphere_verification}

Our sphericity assumption is verified by experiment: latent safe sets can indeed be complex and non-convex, so we designed our Dynamic KL prior to ask the encoder to \emph{reshape} the representation, and we verify that it did (Appendix~\ref{app:necessity}, Fig.~\ref{fig:core_mechanisms}b: a likely fragmented safe set mapped onto a sphere).

\begin{figure}[ht]
    \centering
    \includegraphics[width=\textwidth]{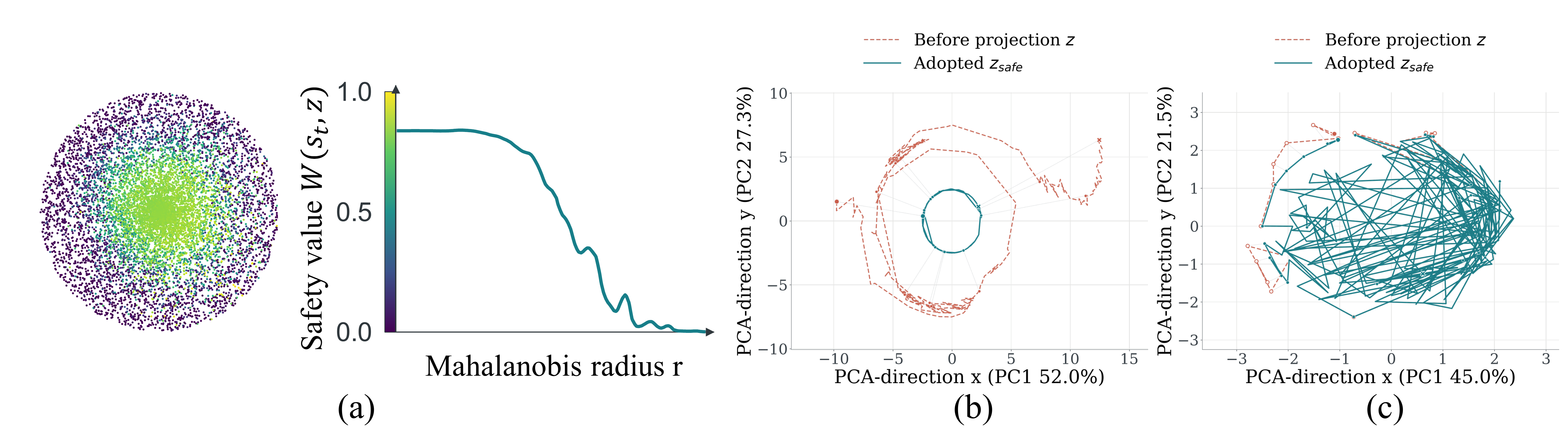}
    \caption{(a) Safety value $W$ vs.\ Mahalanobis radius $r$ over $100$ rollout states. (b,c) PCA traces of $z_t$ under an OOD vs.\ an ID command.}
    \label{fig:sphere_verification}
\end{figure}

For the origin approximation, over $100$ rollout states we searched $\max_{z'} \widehat{W}_\omega(s,z')$ against $\widehat{W}_\omega(s,\mathbf{0})$: the latter is \emph{always} smaller (conservative) by $2.5\%$ on average and $5.2\%$ at worst. Fig.~\ref{fig:sphere_verification}a plots the safety--radius trend: $94\%$ of point pairs are monotonically decreasing (pairwise concordant). A learnt \emph{ellipsoidal} boundary was on par, so we kept the sphere.

\section{Additional OOD Constructions and Baselines}
\label{app:additional_baselines}

Beyond the OOD sets in the main text, we also built OOD sets by trajectory reversal (OOD-Rev), speed rescaling, and axis permutation; all behave like the paper's own sets---CMP raises SR stably at comparable ID precision---by construction the filter is independent of the OOD type.

\textbf{Command clipping:} after extensive design, clipping handles OOD-Geometry and smooths jumps, but no simple rule detects and clips all OOD types (e.g., reversed trajectories (OOD-Rev) in Table~\ref{tab:clipping_baselines}). Recovery policies~\cite{thananjeyan2021recovery} interrupt the task with a non-smooth handover and forfeit accuracy, a fallback when best-effort safety fails, not a replacement.

\begin{table}[ht]
    \centering
    \caption{Each method at its own best, matched budget.}
    \label{tab:clipping_baselines}
    \resizebox{0.8\linewidth}{!}{
    \begin{threeparttable}
        \begin{tabular}{lcccc}
            \toprule
            Method & ID $e_r$ (rad) $\downarrow$ & OOD-Geo SR (\%) $\uparrow$ & OOD-Rev SR (\%) $\uparrow$ & Other OOD Types \\
            \midrule
            UMI-on-Legs & $\mathbf{0.105}$ & $4.7$ & $19.0$ & $\times$ \\
            Command Clipping & $\mathbf{0.107}$ & $\mathbf{90.2}$ & $33.0$ & $\times$ \\
            Rear-augmented & $0.130$ & $87.0$ & $30.2$ & $\times$ \\
            \midrule
            \textbf{CMP} ($R_{safe}=2$) & $\mathbf{0.106}$ & $46.9$ & $\mathbf{65.2}$ & $\checkmark$ \\
            \bottomrule
        \end{tabular}
        \begin{tablenotes}
            \footnotesize
            \item With this large test set, repeated-run s.d.\ for all entries is $\approx 0.002$\,rad for $e_r$, $0.2$\,cm for $e_p$, and $3\%$ for SR or lower.
        \end{tablenotes}
    \end{threeparttable}
    }
\end{table}

\textbf{Why not simply train broader, or use a curriculum:} training with rear-augmented data degraded ID tracking by nearly $30\%$, and switching to another of innumerable OOD types easily erases that gain (like reversed trajectories (OOD-Rev) in Table~\ref{tab:clipping_baselines}). Injecting unmasterable trajectories neither covers OOD nor preserves ID; curricula mitigate but do not solve this (Appendix~\ref{app:necessity}).

\textbf{How the baselines were tuned:} with filters off, backbones have matched parameter counts for fairness. With filters on, Latent Shielding~\cite{nakamura2025generalizing} adds $\ge 1$ encoder-sized forward pass and Neural CBF~\cite{nakamura2025train} $\ge 1$ backward pass; this minimal latency already degrades real-hardware performance, hence CMP's larger lead there. All margins and gains were swept; their Pareto fronts lie below CMP's.

\section{Calibrating the Safety Radius}
\label{app:calibrating_rsafe}

In principle $R_{safe}$ follows from $\delta$, but $\delta$ bounds a \emph{single-step} quantity, the Maximum Probability of Perpetual Safety, hard to relate by hand to whole-trajectory SR. Sweeping $R_{safe}$ takes under five minutes, so we sweep; Fig.~\ref{fig:sim_error_safety} shows mild sensitivity to $R_{safe}$.

\section{Latent Geometry and Failure Analysis}
\label{app:failure_analysis}

We visualize $z_t$ along rollouts in PCA; Fig.~\ref{fig:sphere_verification}(b,c) contrast OOD/ID commands. Under ID, pre-/post-projection traces nearly coincide, whereas the OOD command lands at large radius and is pulled back. ILS also makes $\|z_t\|$ track intent aggressiveness, a free risk signal. Each failure traces to a raw $z$ wrongly judged safe or a projected $z$ still unsafe; a plain CVAE shows more of both.